\documentclass[runningheads]{llncs}
\usepackage[T1]{fontenc}

\usepackage{float}
\usepackage{graphicx}
\usepackage{subcaption}
\usepackage{placeins}
\usepackage{amsmath,amssymb}
\usepackage{hyperref}
\usepackage{booktabs}

\usepackage{listings}
\begin{document}



\title{Accuracy and Robustness of Model Cascades Under Data Perturbations} 

%
%
\author{Pallavi Mitra\inst{1,2} \and
Jai Kushwaha\inst{2} \and
Felix Bie{\ss}mann\inst{2}}
\authorrunning{P. Mitra et al.}
\institute{
AUMOVIO AI Lab, Berlin, Germany\\
\email{pallavi.mitra@aumovio.com}
\and
Berliner Hochschule f\"ur Technik, Berlin, Germany\\
\email{felix.biessmann@bht-berlin.de}
}

\maketitle              

\begin{abstract}
Prediction cascades significantly reduce energy consumption of Artificial Intelligence (AI) models while maintaining high predictive performance. The idea is that easy inputs are routed through a lightweight small model, and difficult uncertain cases are deferred to a larger model. While this design can improve computational efficiency on clean data, its effectiveness depends on the reliability of confidence-based routing. Input degradations, such as static corruptions and sequential perturbations, can shift model confidence and routing decisions. In this paper, we study  confidence-based cascade frameworks for image classification and investigate how such degradations affect their confidence-based deferral behavior. We select a model cascade at the pareto-optimum of accuracy, routing quality, and energy consumption that achieves competitive predictive performance with an up to 10-fold decrease in CO$_2$ emissions. We study the behavior of that model cascade under input corruptions and analyze how the cascade's routing decisions change when the input distribution shifts. Our analysis identifies three failure modes. Static corruptions either (1) break  the routing signal while the large model remains useful, or (2) degrade both models so deferral no longer recovers accuracy. Sequential perturbations reveal a third mode: predictions stabilize but deferral suppresses, yielding stable but unreliable predictions. These findings demonstrate that energy efficient model cascades require evaluation beyond clean accuracy, with explicit attention to routing reliability under distribution shift.

\end{abstract}


\section{Introduction}

The computational cost and energy consumption of the implementation of AI models pose significant environmental challenges~\cite{strubell2019energy,schwartz2020green}. As the scale of the model continues to grow, reducing inference costs has become critical to sustainable AI deployment. Among various efficiency techniques—including pruning~\cite{han2015compression}, knowledge distillation~\cite{hinton2015distilling}, quantization~\cite{jacob2018quantization}, and early-exit architectures~\cite{teerapittayanon2016branchynet}—prediction cascades offer a particularly promising approach by adaptively routing inputs across models of varying complexity. The core idea is intuitive: not all inputs require the same computational effort. Easy samples can be handled by lightweight models, while only difficult cases need expensive large models. The Gatekeeper framework~\cite{rabanser2026gatekeeper} implements this through confidence-based deferral, where a small model ($M_S$) first attempts each prediction and defers uncertain cases to a large model ($M_L$). When a small model $M_S$ is confident, the cascade saves computation; when uncertain, it invests in the capacity of the large model $M_L$. On clean data, this design achieves substantial efficiency gains.

However, real-world deployment rarely matches clean training assumptions. In computer vision applications, for instance, images may suffer from compression artifacts, sensor noise, motion blur, or weather-related degradations. Even small perturbations can shift model predictions and confidence estimates. Although prior work has studied cascade performance on clean data or examined the robustness of individual models, \textbf{the interaction between cascade routing decisions and input degradations remains underexplored}.

This is critical for sustainable AI: cascade efficiency depends entirely on reliable confidence-based routing under distribution shift. Input degradations trigger opposing failures: excessive deferral raises computational cost; suppressed deferral sacrifices accuracy. Therefore, a critical practical question remains unanswered: \textbf{at what level of input degradation does a cascade's efficiency advantage disappear, and can we identify which corruption types pose the greatest threat to routing reliability?}

Accurately characterizing this interaction is essential for estimating the true computational and environmental cost of cascade-based inference in production settings. In this work, we investigate how input degradation -- both static corruptions and sequences of data perturbations -- affects confidence-based deferral in model cascades. Our contributions are:
 
\begin{itemize}
    \item 
     We quantify how severity, type, and temporal structure of input degradation affect model cascades, identifying which static corruption groups and sequential perturbation types most strongly disrupt accuracy, routing reliability, and deferral quality across datasets.
 
    \item 
    We disentangle routing failure from fallback-model degradation, showing that cascade failure can arise from either unreliable confidence-based routing or from collapse of the large model $M_L$ under severe corruption.
    
    \item 
    We show that the relative importance of these failure modes depends on problem complexity: routing calibration becomes the bottleneck in lower-complexity settings, while base-model robustness dominates in higher-complexity settings.
\end{itemize}

\section{Related Work}
In the following sections, we review recent work in model cascades as well as methods focusing on distributional shifts.

\subsection{Prediction Cascades and Efficient Inference}
 
Prediction cascades reduce computational cost by processing inputs through a sequence of models with increasing capacity, allowing early termination when confidence is sufficient. Classical examples include boosting methods such as the Viola-Jones face detector~\cite{viola2001rapid}, which uses a cascade of increasingly complex classifiers to reject non-face regions quickly. Modern approaches extend this concept to deep learning with learned routing strategies~\cite{wang2018skipnet,teerapittayanon2016branchynet}.
In this study we chose a recent model cascading approach, referred to as Gatekeeper~\cite{rabanser2026gatekeeper} which routes each input $x$ using the confidence of the small model $M_S$. If the maximum softmax confidence exceeds an inference threshold $\tau$, the prediction of $M_S$ is accepted; otherwise, the input is deferred to the large model $M_L$:
\begin{equation}
\hat{y} =
\begin{cases}
f_{M_S}(x), & \max_c \sigma(f_{M_S}(x))_c \geq \tau,\\
f_{M_L}(x), & \text{otherwise}.
\end{cases}
\end{equation}
Here, $\tau \in [0,1]$ controls the accept--defer decision at inference time. Gatekeeper fine-tunes $M_S$ with correctness-aware loss
$\mathcal{L}_{\mathrm{GK}}=\alpha\mathcal{L}_{\mathrm{corr}}+(1-\alpha)\mathcal{L}_{\mathrm{incorr}}$,
where $\alpha \in [0,1]$ controls the emphasis between correct and incorrect predictions, $\mathcal{L}_{\mathrm{corr}}$ is cross-entropy on correctly predicted samples, and $\mathcal{L}_{\mathrm{incorr}}$ is KL divergence to a uniform distribution on incorrectly predicted samples.

The framework uses two key metrics to characterize cascade behavior: \textbf{Cascade accuracy} and \textbf{Deferral performance}.
Cascade accuracy simply measures accuracy of the model cascade: 
$\mathrm{Acc}_{\mathrm{casc}}=\frac{1}{N}\sum_{i=1}^{N}\mathbb{1}[\hat{y}_i=y_i]$.
Deferral performance ($s_d$) quantifies routing effectiveness by measuring the normalized area between realized and ideal deferral curves:

\begin{equation}
s_d = \frac{\int_0^1(\mathrm{acc}_{\mathrm{real}}(r)-\mathrm{acc}_{\mathrm{rand}}(r))\,dr}
{\int_0^1(\mathrm{acc}_{\mathrm{ideal}}(r)-\mathrm{acc}_{\mathrm{rand}}(r))\,dr}
\end{equation}

Higher $s_d$ indicates more effective deferral. 
Prior work on cascades has focused primarily on clean, in-distribution data. While efficiency gains are well-documented under these conditions~\cite{wang2018skipnet,teerapittayanon2016branchynet}, the reliability of confidence-based routing under distribution shift remains largely unexplored. Specifically, whether the calibration induced by model cascades, such as Gatekeeper's training loss, remains effective when inputs are corrupted or perturbed is an open question that our work addresses.

\subsection{Distribution Shift, Calibration, and Cascade Routing}
 
Perturbations of input data can degrade model accuracy and confidence calibration~\cite{hendrycks2019robustness,ovadia2019trust}. Hendrycks and Dietterich~\cite{hendrycks2019robustness} introduced standardized static corruption and sequential perturbation benchmarks to measure robustness to common input degradations such as noise, blur, weather effects, and digital artifacts. These benchmarks show that DNNs can suffer substantial performance drops under corrupted inputs, even at moderate severity levels. Beyond accuracy, distribution shift also affects uncertainty estimates: models trained on clean data often become miscalibrated under corrupted inputs~\cite{ovadia2019trust}, while post-hoc calibration methods may not reliably transfer to shifted distributions~\cite{minderer2021revisiting}.

While prior work extensively studies robustness and calibration for individual models, cascade systems introduce an additional failure point: \textbf{routing reliability}. In Gatekeeper cascades, confidence estimates determine whether a sample is accepted by the small model $M_S$ or deferred to the large model $M_L$. Under distribution shift, corrupted confidence signals can therefore alter the accept/defer behavior even when model accuracy alone does not fully explain the failure. Two cascade-specific failure modes can arise: \textit{overconfident acceptance}, where $M_S$ accepts degraded samples that should have been deferred, and \textit{underconfident deferral}, where $M_S$ defers excessively, reducing the efficiency benefit of the cascade. Our work addresses this gap by evaluating how static corruptions and sequential perturbations affect not only cascade accuracy but also deferral performance.



\section{Experimental Setup }

\subsection{Datasets}
We evaluate robustness on two complementary CIFAR benchmarks~\cite{hendrycks2019robustness} to test distinct failure modes: \textbf{Static corruptions} (CIFAR-10-C/100-C) 
apply single-step corruption at fixed severity levels, testing cascade robustness to 19 corruption types at five severity levels, grouped into Noise (Gaussian, shot, impulse), Blur (defocus, glass, motion, zoom), Weather (snow, frost, fog, brightness), and Digital (contrast, elastic, pixelate, JPEG). \textbf{Sequential perturbations} (CIFAR-10-P/CIFAR-100-P) 
present progressive 30-frame degradation sequences, testing cascade stability under sequential gradual input perturbation sequences (motion blur, snow, zoom blur, brightness, fog, Gaussian noise).
Each corruption benchmark contains 10,000 test images per corruption type and severity level.

\subsection{Models and Training}
Following the Gatekeeper formulation~\cite{rabanser2026gatekeeper}, for both CIFAR-10/100, we instantiate $M_S$ as a custom SmallCNN and $M_L$ as ResNet-18. Both models are trained on clean data with standard augmentation. $M_S$ is trained for 50 epochs using Adam with learning rate $10^{-3}$ and weight decay $10^{-4}$. $M_L$ is trained for 200 epochs using SGD with learning rate $0.1$, momentum $0.9$, Nesterov acceleration, weight decay $5\times10^{-4}$, and cosine annealing. We then keep $M_L$ fixed and fine-tune $M_S$ using the Gatekeeper correctness-aware loss for 30 epochs with Adam, learning rate $3\times10^{-4}$, weight decay $10^{-4}$, and cosine annealing. Following the original Gatekeeper 
design~\cite{rabanser2026gatekeeper}, the confidence-based routing threshold $\tau$ is fixed at 0.7. Separate Gatekeeper models are fine-tuned for $\alpha \in \{0.1,0.3,0.5,0.7,0.9\}$.

\subsection{Evaluation Pipeline and Metrics }
After training the Gatekeeper cascade on clean CIFAR-10/100 data, we evaluate it under both clean and corrupted data. We report energy consumption or $CO_2$ emissions as estimated with the codecarbon library~\cite{codecarbon}. For evaluation of performance under corruptions, we apply all corruptions at inference time before the confidence-based routing decision. 
We report accuracy of the small model $M_S$, the large model $M_L$ and the model cascade $GK$, as well as deferral performance ($s_d$) to diagnose failure modes. In addition for the evaluation of sequential perturbations,we report cascade accuracy, deferral rate, and mean cascade flip rate, defined as $\text{mFP-Casc}=\frac{1}{T-1}\sum_{t=2}^{T}\mathbb{1}[\hat{y}_t\neq\hat{y}_{t-1}]$, where $T$ is the sequence length and $\hat{y}_t$ is the cascade prediction at frame $t$. Since sequential perturbations present consecutive frames with progressive degradation, we measure prediction stability by mFP-Casc. 

For grouped results, bars show the mean across corruption or perturbation types within each group, and error bars denote the corresponding standard deviation. These error bars capture variability across degradation types, not variability across independent training seeds.


\section{Result \& Analysis}

\subsection{Model Cascade Performance on Clean Data}
\label{sec:clean_results}

In Table~\ref{tab:pareto}, we report clean-data performance for the Gatekeeper (GK) model cascade on the CIFAR-10 and CIFAR-100 data sets and compare the performance in terms of accuracy, deferral performance $s_d$ and energy consumption for cascades and the large or small models used in the cascade. We observe that for CIFAR-10, the best model cascade with $\alpha=0.9$ yields a competitive accuracy compared to the large model $M_L$ while requiring only 60\% of the energy. For the CIFAR-100 data set we also observe a slight reduction in accuracy but a 10-fold reduction in energy consumption for the model cascade.  All subsequent evaluations use this fixed threshold across all corruption conditions. 
%
\begin{table}[ht]
\centering
\caption{Clean-data accuracy, deferral performance, and estimated CO$_2$ emissions for GK evaluation on CIFAR-10/100. Pareto-optimal settings are shown in bold face.}
\label{tab:pareto}
\small
\setlength{\tabcolsep}{4pt}
\begin{tabular}{l ccc ccc}
\toprule
& \multicolumn{3}{c}{\textbf{CIFAR-10}} 
& \multicolumn{3}{c}{\textbf{CIFAR-100}} \\
Config. \\
& CascAcc$\uparrow$ &  $s_d\uparrow$ & CO$_2\downarrow$ 
& CascAcc$\uparrow$ & $s_d\uparrow$ & CO$_2\downarrow$ \\
\midrule
$M_S$ only
  & 0.752  & 0.674 & 0.029
  & 0.357  & 0.496 & 0.042 \\
GK $\alpha{=}0.1$
  & 0.584  & 0.861 & 0.039
  & 0.687  & 0.986 & 0.030 \\
GK $\alpha{=}0.3$
  & 0.878  & 0.788 & 0.042
  & 0.708  & 0.804 & 0.029 \\
GK $\alpha{=}0.5$
  & 0.900  & 0.743 & 0.041
  & 0.729  & 0.748 & 0.029 \\
GK $\alpha{=}0.7$
  & 0.910  & 0.731 & 0.029
  & 0.737  & 0.714 & 0.029 \\
\textbf{GK $\alpha{=}0.9$}
  & \textbf{0.922}  & 0.686 & \textbf{0.029}
  & \textbf{0.744}  & 0.667 & \textbf{0.029} \\
$M_L$ only
  & 0.941 & -- & 0.0476
  & 0.771  & -- & 0.295 \\  
\bottomrule
\end{tabular}

\footnotesize{\textit{Note:} $s_d$ is undefined for $M_L$-only since there is no deferral.}
\end{table}

\subsection{Model Cascade Performance on Corrupted Data}
\label{sec:corruption}
\begin{figure*}[!t]
    \centering
    \setlength{\tabcolsep}{2pt}
    \renewcommand{\arraystretch}{0.2}

    \begin{tabular}{cc}
        \includegraphics[width=0.47\textwidth]{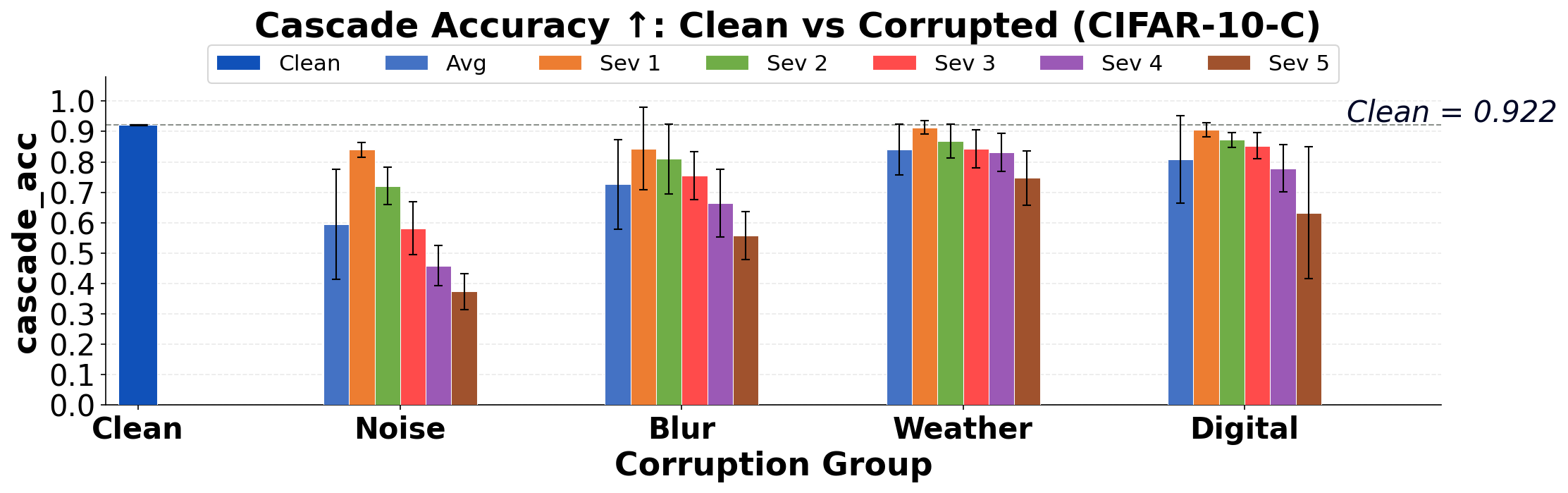} &
        \includegraphics[width=0.47\textwidth]{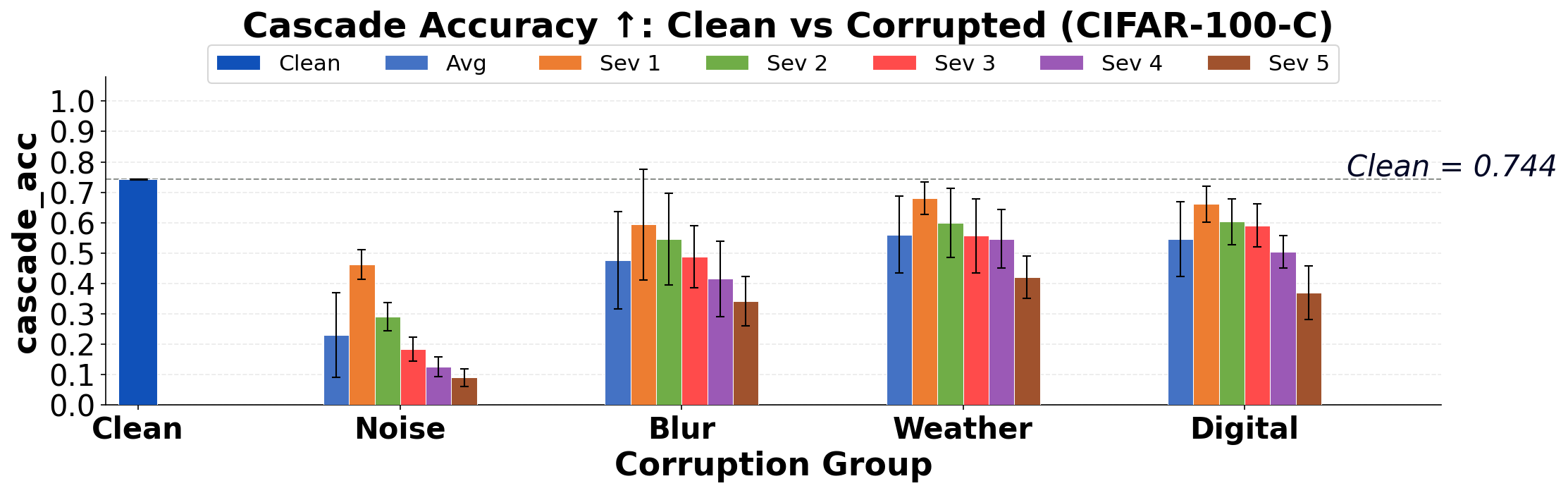} \\[-0.65em]

        \includegraphics[width=0.47\textwidth]{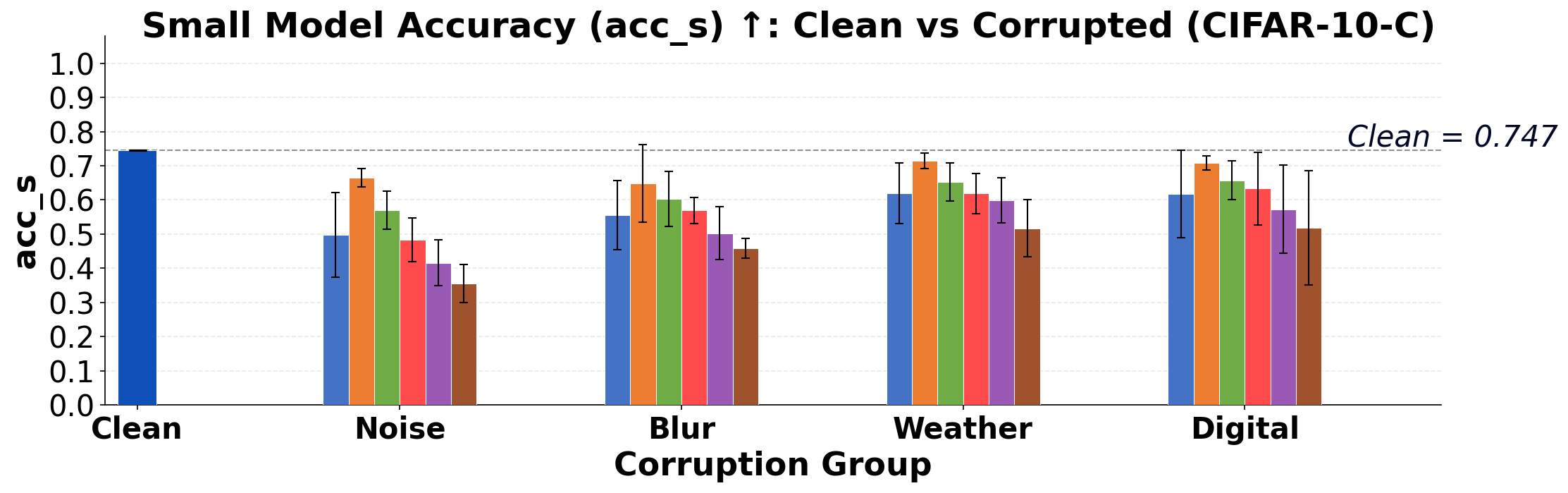} &
        \includegraphics[width=0.47\textwidth]{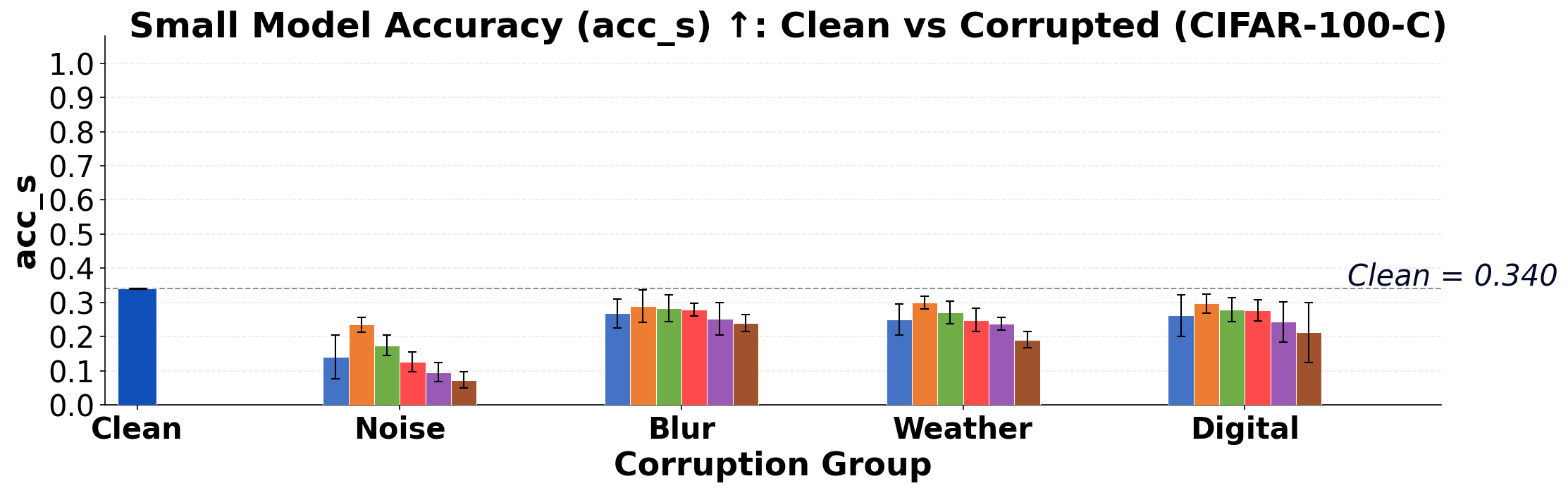} \\[-0.65em]

        \includegraphics[width=0.47\textwidth]{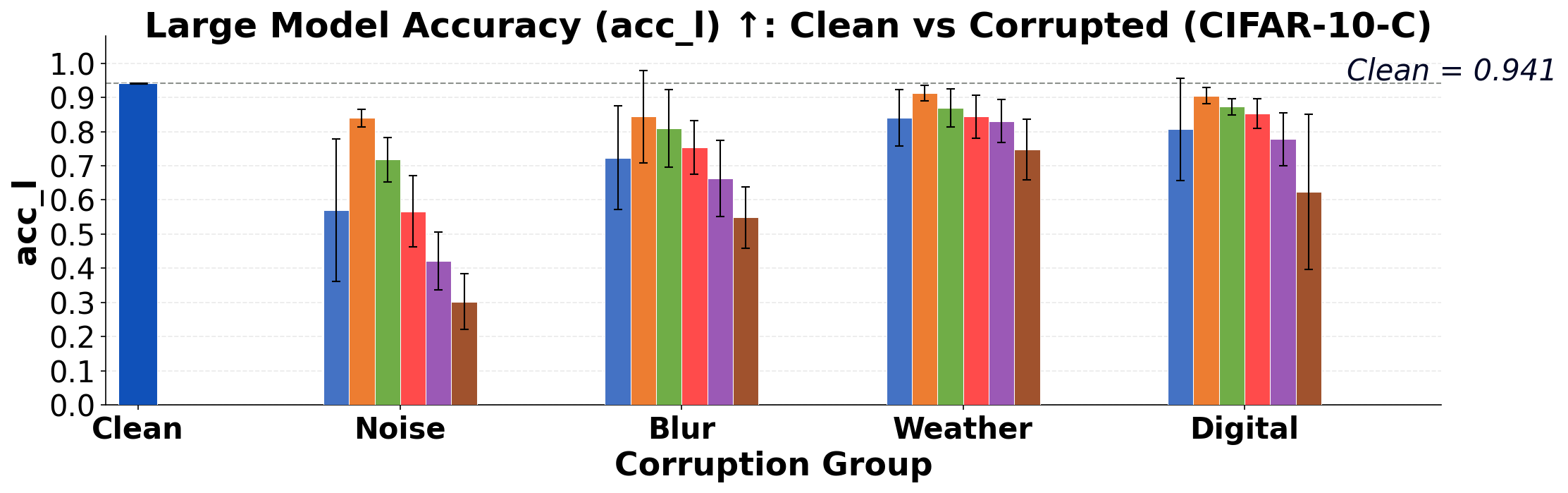} &
        \includegraphics[width=0.47\textwidth]{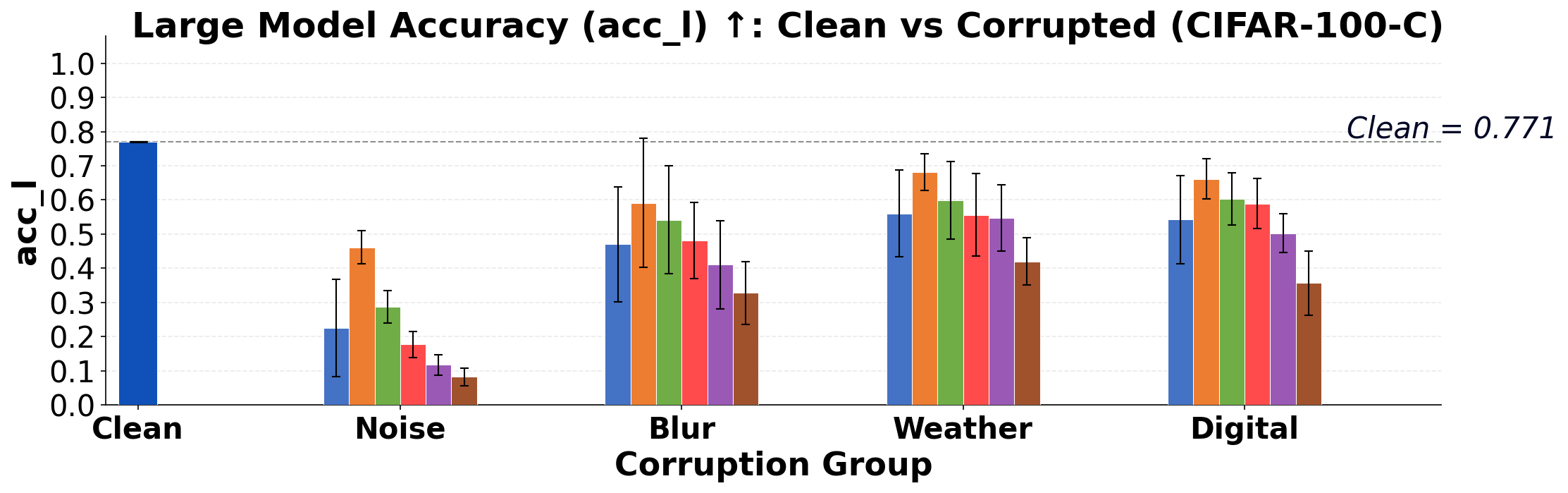} \\[-0.65em]

        \includegraphics[width=0.47\textwidth]{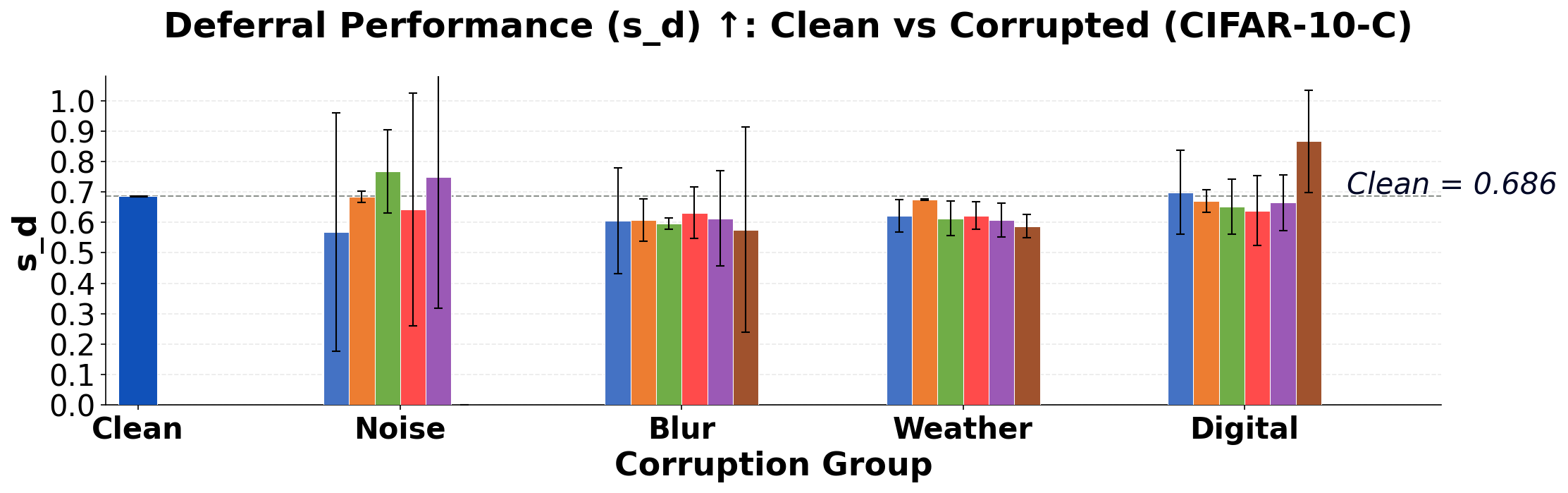} &
        \includegraphics[width=0.47\textwidth]{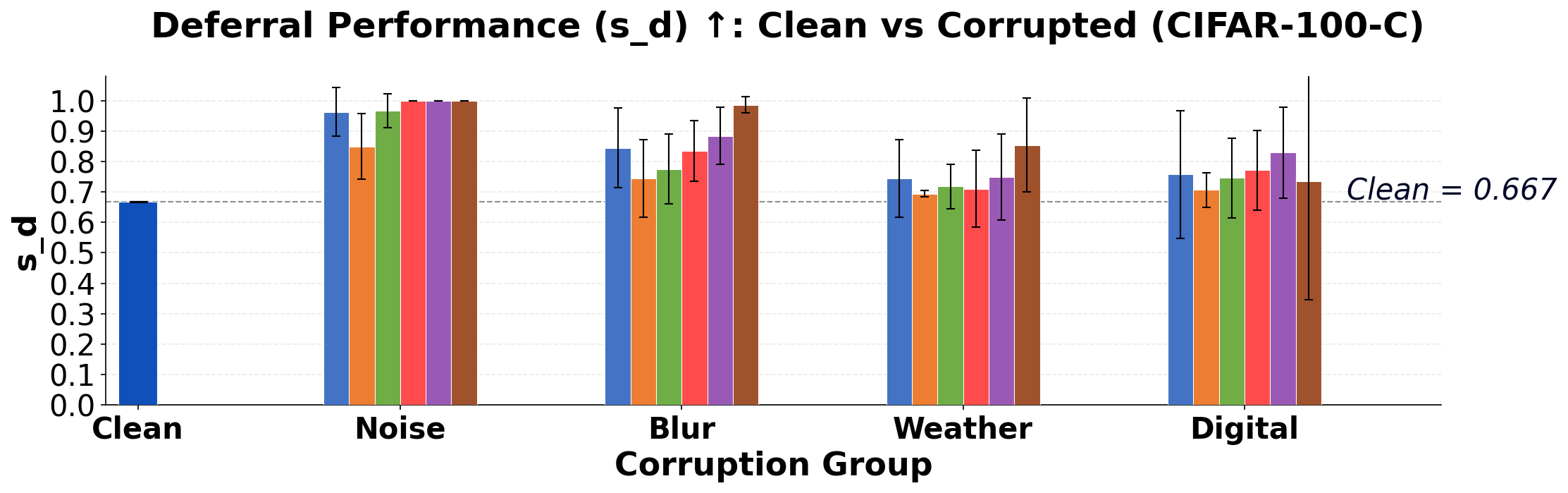} 
        
    \end{tabular}

    \caption{Corruption robustness on CIFAR-10-C (left) and CIFAR-100-C (right). For the CIFAR-10-C data the large model $M_L$ remains accurate but the lower deferral score $s_d$ indicates routing failure. For the CIFAR-100-C data even the large model $M_L$ performs poorly under Noise, limiting the model cascade performance. } 
    \label{fig:corruption_results}
 
\end{figure*}

%
%
The results in \autoref{fig:corruption_results} demonstrate that the small model $M_S$, the large model $M_L$ as well as the model cascade perform less accurately when data is corrupted. Increasing levels of corruption severity lead to decreasing levels of accuracy. On CIFAR-10-C, cascade accuracy drops from $0.922$ on clean data to $\approx 0.38$--$0.58$ under Noise across severity levels, while Weather remains comparatively robust, staying around $0.75$--$0.90$. In contrast, CIFAR-100-C shows a much sharper degradation: cascade accuracy drops from $0.744$ on clean data to approximately $0.46$ under Noise at severity~1 and to $\approx 0.10$ at severity~5. Even the most robust Weather group remains only around $0.42$--$0.68$ across severities. Overall the impact of corruptions on the 10-class problem CIFAR-10 is much less pronounced than on the task with larger label-set cardinality CIFAR-100. The clean-to-worst-severity drop is larger on CIFAR-100-C: under Noise, accuracy decreases by $\approx 0.64$ ($0.744 \rightarrow 0.10$) compared with $\approx 0.54$ on CIFAR-10-C ($0.922 \rightarrow 0.38$); under Weather, the drop is $\approx 0.32$ (CIFAR-100) versus $\approx 0.17$ (CIFAR-10).
Comparing the impact of different corruptions on model cascade accuracy we observe a consistent ranking across datasets: \textbf{Noise$\gg$ Blur $>$ Weather $\approx$ Digital}. Weather corruptions have the least impact on accuracy. Interestingly, the severity level at which the cascade loses its efficiency advantage is dataset-dependent. On CIFAR-10-C, Weather and Digital remain above $0.75$ accuracy through severity~3, whereas on CIFAR-100-C no corruption group remains above $0.60$ beyond severity~3.

In order to investigate whether this difference in corruption impact on model cascade performance is due to the model cascade's deferral performance or due to the task difficulty itself we inspect the deferral ratios $s_d$ (~\autoref{fig:corruption_results}, bottom row). On CIFAR-10-C, the large model $M_L$ retains a clear accuracy advantage over $M_S$ under corruption, enabling deferral to recover useful accuracy when routing remains reliable. For example, under Noise at severity~1: $\mathrm{acc}_L = 0.84$ vs. $\mathrm{acc}_S = 0.67$ (27 pp gap). In contrast, on CIFAR-100-C, severe Noise corruption degrades both the large model $M_L$ and the small model $M_S$ to similarly low accuracy ($<0.10$), eliminating the utility of the large model as a fallback.

These results suggest that corruptions impact the performance of the model cascade in different ways: For CIFAR-10-C we observe that the large model $M_L$ remains substantially more accurate than the small model $M_S$ under corruptions (e.g. 0.84 vs 0.67 under Noise-1), but the degraded confidence signals (s$_d$: 0.686→0.35 for Noise) prevent the cascade from exploiting this high accuracy of the large model $M_L$. Due to poor deferral performance, the cascade loses $\approx$19pp of recoverable accuracy.

In contrast for CIFAR-100-C especially high severity corruptions degrade both the accuracy of the small model $M_S$ as well as the large model $M_L$, eliminating the large model's advantage. These results suggest that it is rather the task difficulty than the deferral performance of the model cascade that leads to the lower cascade accuracy. If the large model $M_L$ is not able to correctly identify samples, the model cascade will not be able to compensate for that.

\subsection{Model Cascade Performance on Sequential Corruptions}
\label{sec:perturbation}
\begin{figure*}[!t]
    \centering
    \begin{tabular}{cc}
        \includegraphics[width=0.47\textwidth]{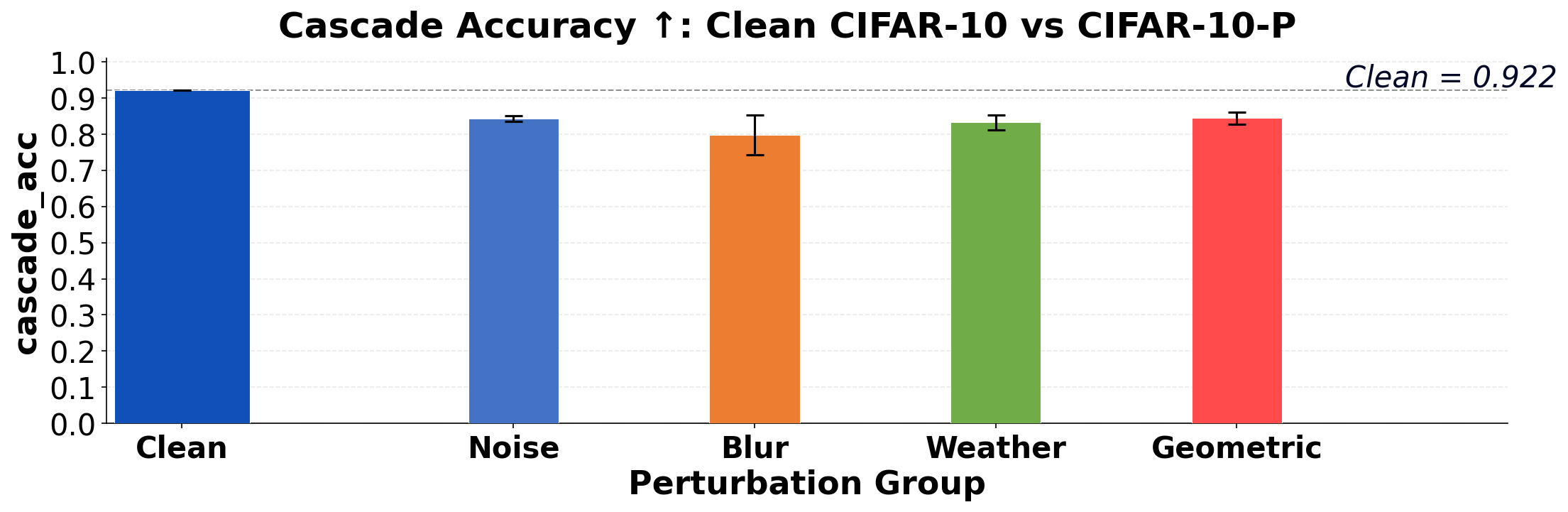} &
        \includegraphics[width=0.47\textwidth]{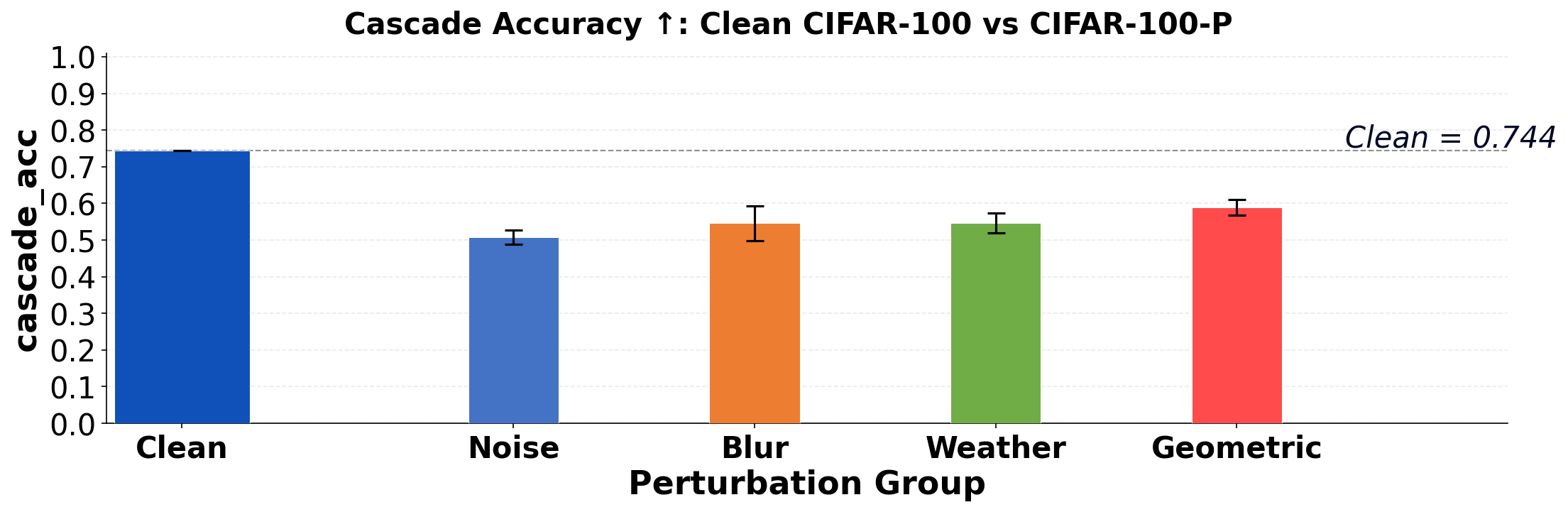}\\[-0.90em]

        \includegraphics[width=0.47\textwidth]{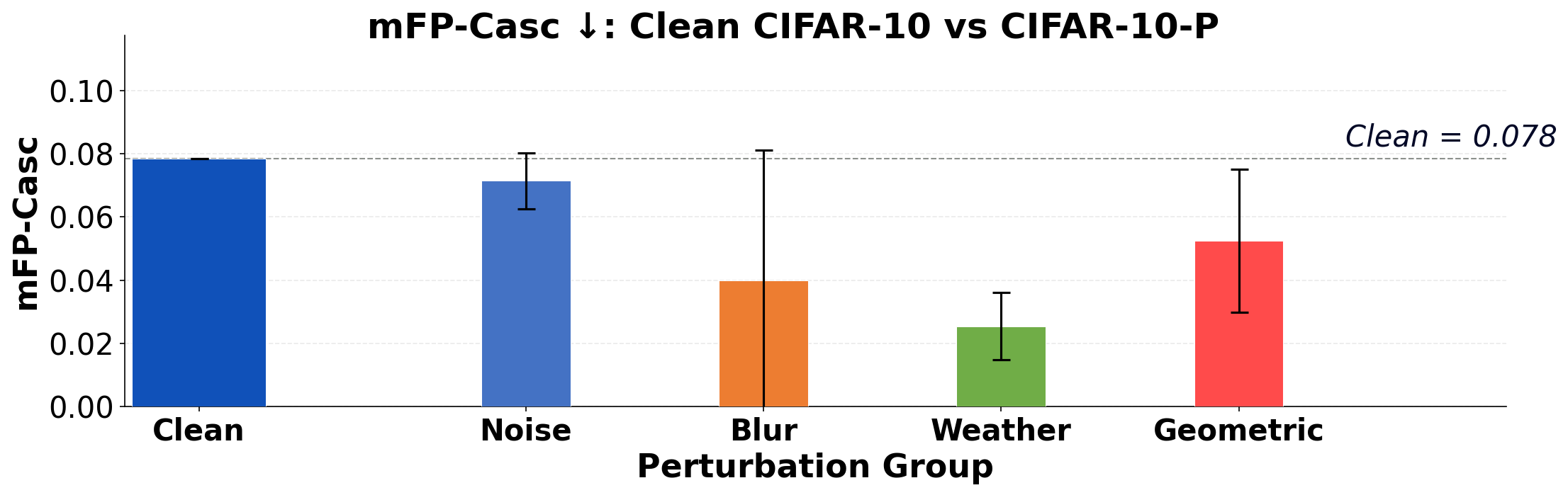} &
        \includegraphics[width=0.47\textwidth]{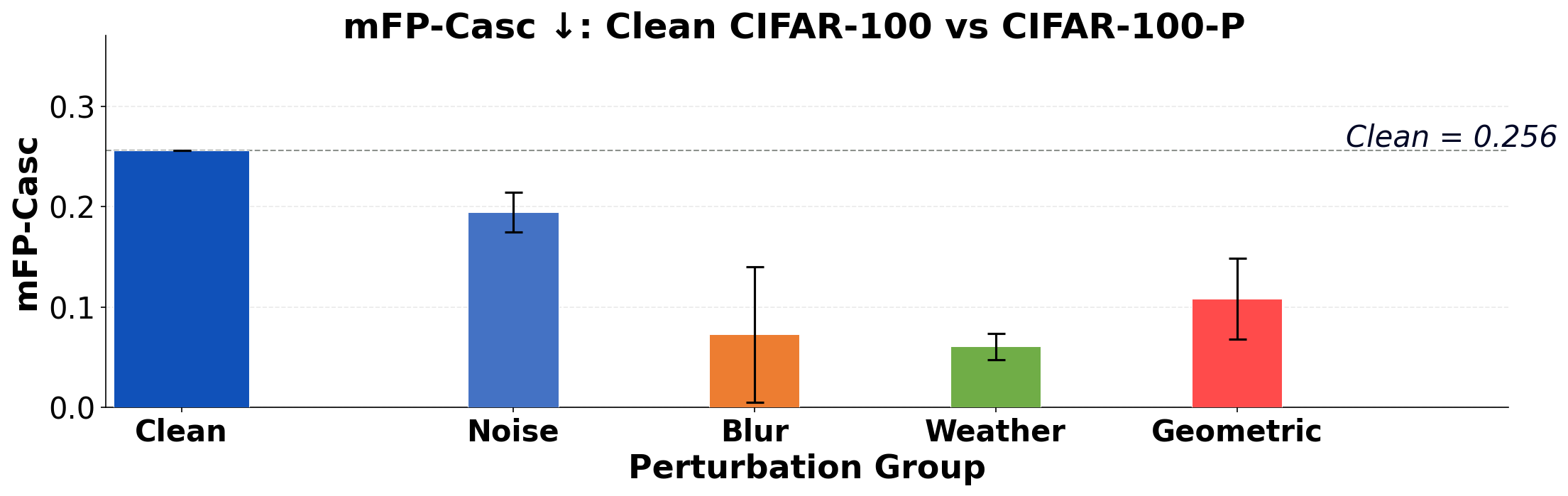} \\[-0.90em]

        \includegraphics[width=0.47\textwidth]{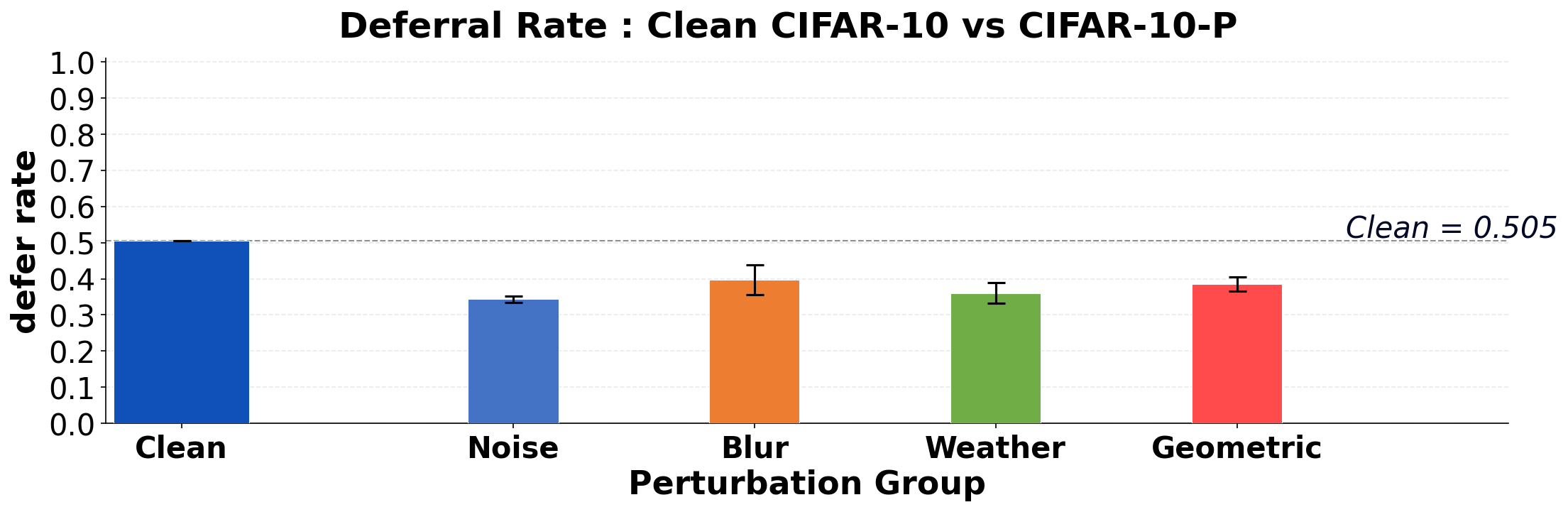} &
        \includegraphics[width=0.47\textwidth]{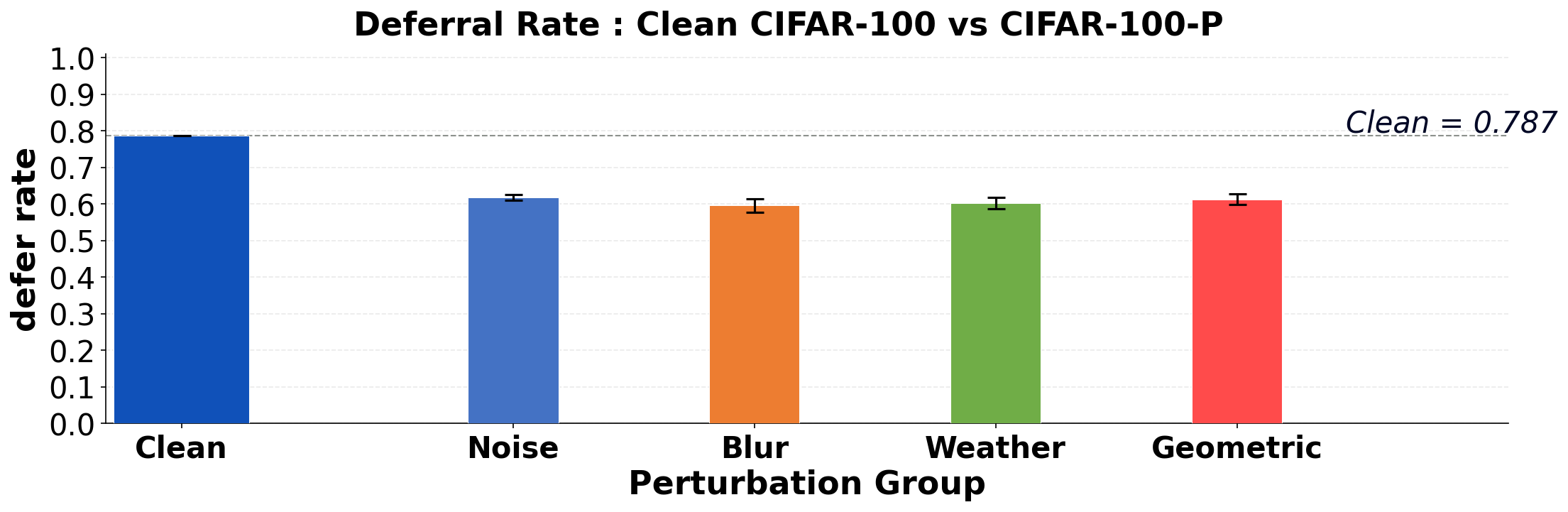}
        
    \end{tabular}
    \caption{Robustness under perturbation sequences on CIFAR-10-P (left) and CIFAR-100-P (right). Perturbations reduce cascade accuracy and mean flip rate, but also suppress deferral to the large model $M_L$. Thus, lower prediction instability can be misleading: the cascade appears stable while relying more on the degraded small model $M_S$.}
    \label{fig:perturbation_results}
\end{figure*}

Evaluating the impact of sequential data perturbations in \autoref{fig:perturbation_results} we observe that progressive perturbations in the CIFAR-10-P and CIFAR-100-P tasks reduce the cascade accuracy, but less severely than static corruptions. On CIFAR-10-P, accuracy drops from 0.922 to 0.80--0.85; on CIFAR-100-P, from 0.744 to 0.51--0.59. Unlike static corruptions where Noise 
dominates, Blur emerges as the most damaging perturbation on CIFAR-10-P, indicating that gradual, frame-by-frame degradation affects confidence-based routing differently than single-step corruptions.

Examining the mean cascade flip rate (mFP-Casc), measuring how often the deferral decision flips with progressive perturbations, reveals a counterintuitive pattern: perturbations reduce prediction instability across all groups. Weather yields the 
largest reduction (67\% on CIFAR-10-P; 76\% on CIFAR-100-P), while Noise yields the smallest (9\% and 24\%, respectively).
However, this apparent stability is deceptive. The deferral rate also decreases under perturbation, from 0.505 to 0.34--0.39 on CIFAR-10-P and from 0.787 to 0.59--0.62 on CIFAR-100-P. Fewer samples are routed to the large model $M_L$, and the cascade relies more heavily on the degraded small model $M_S$. Combined with the observed accuracy drop, the lower flip rate does not indicate improved robustness, but rather reflects a suppressed fallback to the large model $M_L$.
These results suggest that perturbations expose a distinct cascade failure mode: progressive degradation suppresses deferral decisions, producing stable but less reliable cascade predictions. For cascaded systems under temporal degradation, robustness must be evaluated jointly through accuracy, prediction stability, and deferral behavior.

\section{Conclusion and Future Work}


We investigated the impact of data corruptions on the performance of model cascades. In line with previous work we find that on clean data model cascades can achieve competitive predictive performance while obtaining an up to 10-fold reduction in energy consumption. However, our results demonstrate that input degradation affects not only cascade accuracy but also the confidence-based routing decisions that determine whether efficient inference remains reliable. The failure mode of model cascades differs across datasets: In the CIFAR-10-C task we observe mainly routing failures, where the large model $M_L$ retains its high accuracy under data corruptions but the corrupted confidence signals of the smaller models limit effective deferral. In the case of the CIFAR-100-C task we see that also the larger model exhibits decreased performance under data corruptions, limiting its utility as a fallback option for deferral of difficult samples.

Overall our results indicate that model cascades can be a viable alternative to large or small models as they offer a convenient per-sample tradeoff between predictive performance and energy consumption. Using them responsibly requires better understanding of their behaviour under data set shifts. We observe that model cascades can fail under corruption through either routing breakdown (when the large model remains accurate but confidence signals corrupt) or model collapse (when distribution shift degrades both models). In our experiments, the label set cardinality is indicative of the failure mode: CIFAR-10's lower number of classes preserves $M_L$ robustness, exposing routing fragility; CIFAR-100's higher label set cardinality breaks both models, revealing fundamental model brittleness.

Future work will investigate shift-aware Gatekeeper mechanisms that adapt the inference threshold under degraded inputs, rather than using a fixed clean-data configuration. Another direction is to incorporate uncertainty calibration or corruption-aware confidence correction into the routing decision, so that degraded samples are not incorrectly accepted by $M_S$. Finally, extending this analysis to larger-scale datasets, additional model families, and real deployment measurements of latency, energy, and CO$_2$ emissions would provide a more complete picture of the robustness--efficiency trade-off in cascade-based inference.

\section{Acknowledgement}
\label{sec:acknowledgement}

This work was funded by the German Federal Ministry for Economic Affairs and Energy within the project ``Safe AI Engineering -- Sicherheitsargumentation bef\"ahigendes AI Engineering \"uber den gesamten Lebenszyklus einer KI-Funktion''. This research was also supported by the German Research Foundation (DFG), project number 528483508 -- FIP 12. The authors thank the project partners for the successful cooperation.

\FloatBarrier
\bibliographystyle{splncs04}
\bibliography{sample-ceur}

@misc{codecarbon,
  author       = {Benoit Courty and
                  Victor Schmidt and
                  Goyal-Kamal and
                  MarionCoutarel and
                  Boris Feld and
                  J{\'e}r{\'e}my Lecourt and
                  LiamConnell and
                  SabAmine and
                  inimaz and
                  supatomic and
                  Mathilde L{\'e}val and
                  Luis Blanche and
                  Alexis Cruveiller and
                  ouminasara and
                  Franklin Zhao and
                  Aditya Joshi and
                  Alexis Bogroff and
                  Amine Saboni and
                  Hugues de Lavoreille and
                  Niko Laskaris and
                  Edoardo Abati and
                  Douglas Blank and
                  Ziyao Wang and
                  Armin Catovic and
                  alencon and
                  Micha{\l} St{\k e}ch{\l}y and
                  Christian Bauer and
                  Lucas-Otavio and
                  JPW and
                  MinervaBooks},
  title        = {mlco2/codecarbon: v2.4.1},
  month        = may,
  year         = {2024},
  publisher    = {Zenodo},
  version      = {v2.4.1},
  doi          = {10.5281/zenodo.11171501},
  url          = {https://doi.org/10.5281/zenodo.11171501}
}

@article{strubell2019energy,
  title={Energy and Policy Considerations for Deep Learning in NLP},
  author={Strubell, Emma and Ganesh, Ananya and McCallum, Andrew},
  journal={Proceedings of ACL},
  pages={3645--3650},
  year={2019}
}

@article{schwartz2020green,
  title={Green AI},
  author={Schwartz, Roy and Dodge, Jesse and Smith, Noah A and Etzioni, Oren},
  journal={Communications of the ACM},
  volume={63},
  number={12},
  pages={54--63},
  year={2020}
}

@inproceedings{han2015compression,
  title={Learning both Weights and Connections for Efficient Neural Networks},
  author={Han, Song and Pool, Jeff and Tran, John and Dally, William J},
  booktitle={Advances in Neural Information Processing Systems (NeurIPS)},
  pages={1135--1143},
  year={2015}
}

@article{hinton2015distilling,
  title={Distilling the Knowledge in a Neural Network},
  author={Hinton, Geoffrey and Vinyals, Oriol and Dean, Jeff},
  journal={arXiv preprint arXiv:1503.02531},
  year={2015}
}

@inproceedings{jacob2018quantization,
  title={Quantization and Training of Neural Networks for Efficient Integer-Arithmetic-Only Inference},
  author={Jacob, Benoit and Kligys, Skirmantas and Chen, Bo and Zhu, Menglong and Tang, Matthew and Howard, Andrew and Adam, Hartwig and Kalenichenko, Dmitry},
  booktitle={Proceedings of the IEEE Conference on Computer Vision and Pattern Recognition (CVPR)},
  pages={2704--2713},
  year={2018}
}

@inproceedings{teerapittayanon2016branchynet,
  title={BranchyNet: Fast Inference via Early Exiting from Deep Neural Networks},
  author={Teerapittayanon, Surat and McDanel, Bradley and Kung, Hsiang-Tsung},
  booktitle={International Conference on Pattern Recognition (ICPR)},
  pages={2464--2469},
  year={2016}
}

@inproceedings{viola2001rapid,
  title={Rapid Object Detection using a Boosted Cascade of Simple Features},
  author={Viola, Paul and Jones, Michael},
  booktitle={Proceedings of the IEEE Conference on Computer Vision and Pattern Recognition (CVPR)},
  volume={1},
  pages={I--I},
  year={2001}
}

@inproceedings{wang2018skipnet,
  title={SkipNet: Learning Dynamic Routing in Convolutional Networks},
  author={Wang, Xin and Yu, Fisher and Dou, Zi-Yi and Darrell, Trevor and Gonzalez, Joseph E},
  booktitle={European Conference on Computer Vision (ECCV)},
  pages={409--424},
  year={2018}
}

@article{rabanser2026gatekeeper,
  title={Gatekeeper: Improving model cascades through confidence tuning},
  author={Rabanser, Stephan and Rauschmayr, Nathalie and Kulshrestha, Achin and Poklukar, Petra and Jitkrittum, Wittawat and Augenstein, Sean and Wang, Congchao and Tombari, Federico},
  journal={Advances in Neural Information Processing Systems},
  volume={38},
  pages={19518--19547},
  year={2026}
}

@inproceedings{hendrycks2019robustness,
  title={Benchmarking Neural Network Robustness to Common Corruptions and Perturbations},
  author={Hendrycks, Dan and Dietterich, Thomas},
  booktitle={International Conference on Learning Representations (ICLR)},
  year={2019}
}

@inproceedings{ovadia2019trust,
  title={Can You Trust Your Model's Uncertainty? Evaluating Predictive Uncertainty Under Dataset Shift},
  author={Ovadia, Yaniv and Fertig, Emily and Ren, Jie and Nado, Zachary and Sculley, D and Nowozin, Sebastian and Dillon, Joshua V and Lakshminarayanan, Balaji and Snoek, Jasper},
  booktitle={Advances in Neural Information Processing Systems (NeurIPS)},
  pages={13991--14002},
  year={2019}
}

@inproceedings{minderer2021revisiting,
  title={Revisiting the Calibration of Modern Neural Networks},
  author={Minderer, Matthias and Djolonga, Josip and Romijnders, Rob and Hubis, Frances and Zhai, Xiaohua and Houlsby, Neil and Tran, Dustin and Lucic, Mario},
  booktitle={Advances in Neural Information Processing Systems (NeurIPS)},
  pages={15682--15694},
  year={2021}
}

\appendix

\end{document}